\documentclass[letterpaper, 10 pt, conference]{ieeeconf}  

\IEEEoverridecommandlockouts                              

\usepackage{booktabs}
\usepackage{amsmath}
\usepackage{amsfonts}
\usepackage[ruled,linesnumbered]{algorithm2e}
\usepackage{graphicx}
\usepackage{bm}
\usepackage{subcaption}
\usepackage{xcolor}
\usepackage[T1]{fontenc}
\usepackage{lmodern}
\usepackage{tcolorbox}
\usepackage{booktabs} 

\newcommand{\tonghe}[1]{\textcolor{black}{#1}}

\title{\LARGE \bf
\tonghe{
Think on your feet: 
Seamless Transition between Human-like Locomotion 
in Response to Changing Commands
}
}

\begin{document}    
\author{Huaxing Huang~$^{*\ 1}$, Wenhao Cui~$^{*\ 1}$ Tonghe Zhang~$^{*\ 2}$,
\\
Shengtao Li~$^{1}$, Jinchao Han~$^{1}$, Bangyu Qin~$^{1}$, Tianchu Zhang~$^{1}$, 
\\
Liang Zheng~$^{1}$, Ziyang Tang~$^{1}$, Chenxu Hu~$^{1}$
\\ Shipu Zhang~$^{1}$, Zheyuan Jiang~$^{\dagger,1}$
\thanks{*Denotes equal contribution.}
\thanks{$^{1}$Noetix Robotics. $^{2}$Tsinghua University.}
\thanks{$^{\dagger}$ Correspondence to: {\tt\small merlin.jiang@noetixrobotics.com}}
}
\maketitle
\begin{figure*}[ht]
    \centering
    \vspace{-1mm}
    \includegraphics[width=0.95\textwidth,height=0.42\textwidth]{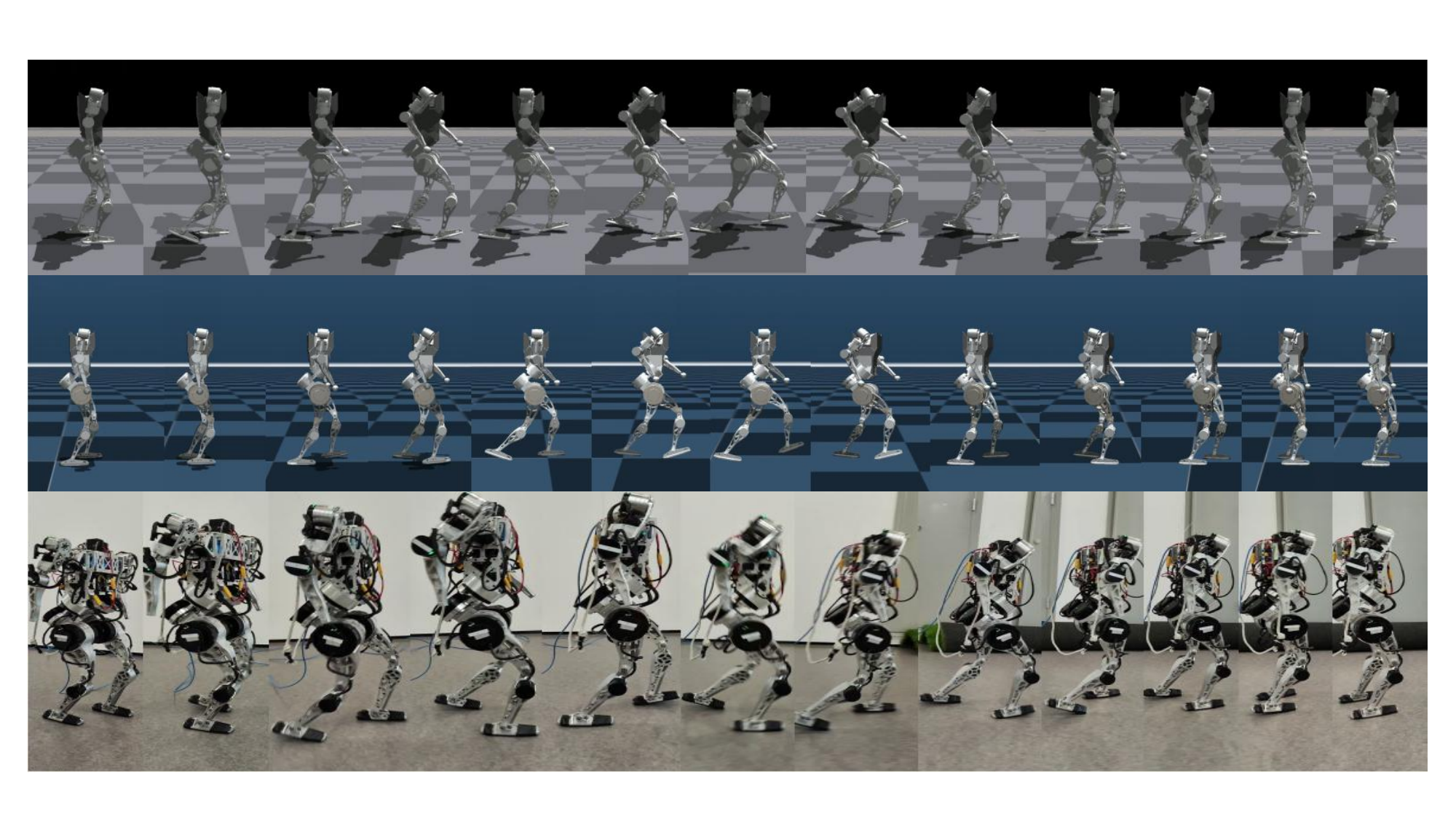}
    \vspace{-3mm}
    \caption{ \small 
        Comprehensive demonstration of Noetix robot N1's 
    locomotion skills learnt from the proposed method. The robot exhibits seamless and continuous transfer between highly human-like motion sets, accelerating from walking to running then coming to a full stop. Top to down: performance in simulator Isaac gym, Mujoco, and the real world. Parameters of the PD controller and the driving frequency are consistent in the three groups.
    }
    \label{fig:real}
    \vspace{-5mm}
\end{figure*}

\begin{abstract}

    \tonghe{ 
    While it is relatively easier to train humanoid robots to mimic specific locomotion skills, 
    it is more challenging to learn from various motions and adhere to continuously changing commands. 
    These robots must accurately track motion instructions, 
    seamlessly transition between a variety of movements,} and master intermediate motions not present in their reference data. 
    In this work, we propose a novel approach that integrates human-like motion transfer with precise velocity tracking by a series of improvements to classical imitation learning. 
    To enhance generalization, we employ the Wasserstein divergence criterion (WGAN-div). 
    Furthermore, a Hybrid Internal Model provides structured estimates of hidden states and velocity to enhance mobile stability and environment adaptability, while a curiosity bonus fosters exploration. 
    Our comprehensive method promises highly human-like locomotion that adapts to varying velocity requirements, 
    direct generalization to unseen motions and multitasking, as well as 
    zero-shot transfer to the simulator and the real world across different terrains.
    These advancements are validated through simulations across various robot models and extensive real-world experiments.
    
\end{abstract}

\section{INTRODUCTION}
Humanoid robots possess great potential to mimic human behaviors, making them ideal for adapting to human-centric environments like factories and homes. However, unlike quadrupedal robots, humanoid robots struggle to learn locomotion skills effectively due to challenges such as a higher center of gravity, increased degrees of freedom, and larger body size.

Reinforcement learning (RL) has proven effective in teaching humanoids basic locomotion skills with minimal prior knowledge. However, exploring under weak reward signals can lead to unnatural gaits, resulting in high energy costs and mechanical wear that impede real-world deployment.

Imitation learning (IL) methods, such as Adversarial Motion Prior (AMP)~\cite{Peng_2021}, are promising for generating fluid human-like motions by mimicking human demonstrations. Nonetheless, their application is limited by their supervised learning nature, leading to restricted generalization and a heavy reliance on high-quality expert data, which is expensive for robotics research~\cite{ARGALL2009469}.

These limitations hinder classical RL and IL algorithms from enabling humanoids to adopt fluid and versatile motion strategies like humans. This raises the question:

\textbf{“Can we achieve seamless transitions between various human-like motions--such as walking, running, and spinning--that adapt to continuously changing velocity commands?'}

In this work, we affirm this question by proposing a novel humanoid locomotion learning algorithm. We integrate the human-mimicking capabilities of IL methods within an advanced RL framework while ensuring strong adherence to input velocity commands. Robots trained using our method adapt to diverse motion instructions while consistently exhibiting human-like movements, even in intermediate stages not covered in the expert data.

Our contributions include:
\begin{itemize}
    \item \textbf{Novel architecture}: We developed a locomotion learning framework combining a Hybrid Internal Model with a WGAN-div module, enabling accurate human motion mimicry and command tracking. 
    \tonghe{
    }
    \item \textbf{Enhanced generalization}: Incorporating a curiosity HashNet and Wasserstein divergence loss functions allows humanoids to master unseen intermediate movements during transitions, significantly advancing imitation-learning techniques. 
    \item \textbf{Extensive empirical evidence}: Our experiments in both simulation and real-world settings demonstrate strong adaptability to various movement commands while maintaining natural, human-like motions.
\end{itemize}

\section{Related Work}\label{sec:Related Work}
\subsection{Reinforcement Learning for Legged Locomotion}
Reinforcement learning has been extensively studied in robotic 
locomotion~\cite{hwangbo2019learning, miki2022learning}.
For instance, \cite{rudin2022learning} describes a setup that rapidly generates policies for robotic tasks using parallelism on a single GPU. Additionally, \cite{nahrendra2023dreamwaq} proposes an estimator that encodes environmental parameters through proprioceptive state histories.

While quadruped locomotion excels in navigating complex terrains, 
humanoid locomotion faces unique challenges due to 
higher degrees of freedom and the need for dynamic balance \cite{article4}.
\cite{gu2024humanoidgym} developed an RL-based humanoid 
locomotion framework using the Legged Gym platform.
 \cite{li2024reinforcement} created a framework 
for training robust controllers for walking, 
running, and jumping skills. 

\begin{figure}[t]
        \centering
        \includegraphics[width=0.48\textwidth]{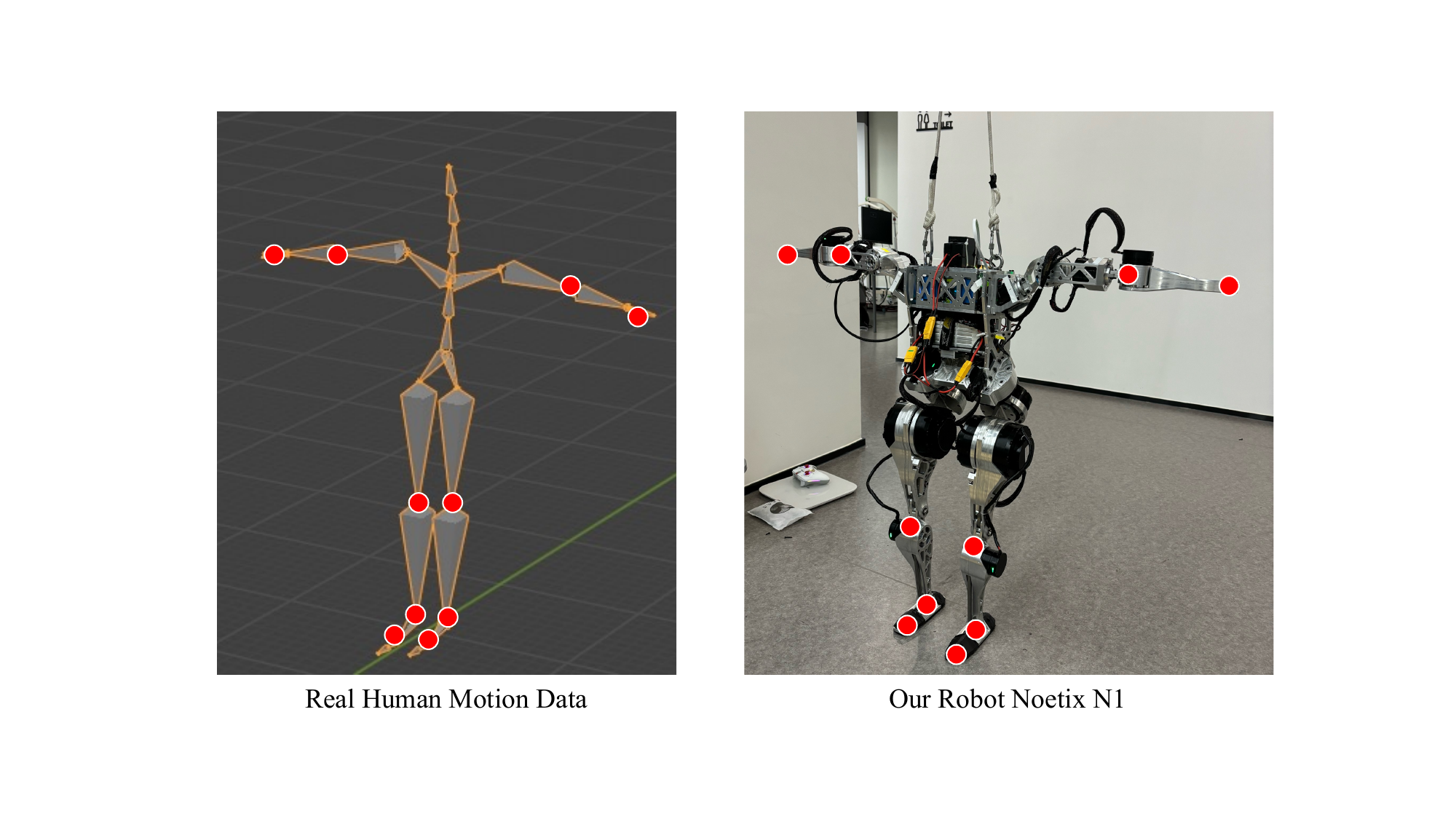}
        \caption{ \small 
        Illustration of motion re-targeting from expert data (Left) to 
        our humanoid robot ``Noetix N1''(Right). 
        N1 weights 23 kg and is of height ass 0.95 m, with 18 DoFs in total (four on each arm and five on each leg).
        }
        \label{fig:retarget}
        \vspace{-6mm}
\end{figure}

\subsection{Motion Imitation}
\tonghe{ 
}
Imitation Learning (IL) effectively tracks joint trajectories and extracts gait features but can struggle with discontinuities between locomotion patterns.
Generative Adversarial Imitation Learning (GAIL)~\cite{ho2016generative} addresses these continuity challenges. 
Innovations like Adversarial Motion Priors (AMP) enhance the generation of realistic motions from unstructured datasets, 
avoiding manual motion design constraints. 
These advancements support agile movements in quadrupedal and humanoid robots using refined 
IL techniques \cite{tang2024humanmimic,wu2023learning, zhang2024whole}.

Building on these methodologies, our research proposes a framework that enhances robotic adaptability and performance in real-world environments.

\section{Problem Setup}\label{sec:setup}
\vspace{-1mm}
We model humanoid locomotion control as optimizing a partially observable Markov decision process $\mathcal{P} = (\mathcal{S}, \mathcal{O}, \mathcal{A}, p, r, \gamma)$, where state, observation, and action are denoted as $\mathbf{s} \in \mathcal{S}$, $\mathbf{o} \in \mathcal{O}$, and $\mathbf{a} \in \mathcal{A}$. The state transition probability is defined as $p(\mathbf{s}_{t+1}|\mathbf{s}_t, \mathbf{a}_t)$. 
The policy $\mathbf{\pi}$ selects actions based on 
$H$ steps of historical observations 
$\mathbf{a}_t \sim \mathbf{\pi}(\cdot | \mathbf{o}_{t}^H)$. 
The reward function is $r_t = r(\mathbf{s}_t, \mathbf{a}_t)$ with a discount factor $\gamma \in [0, 1)$, while 
the objective is to maximize cumulative discounted rewards:\ 
$
J(\pi) = \mathbb{E}\left[ \sum_{t=0}^{+\infty} \gamma^t r(\mathbf{s}_t, \mathbf{a}_t) \right].
$
Table~\ref{tab:o_s_def} summarizes the construction and physical meaning of the spaces. 
\begin{table}[htbp]\centering
    \caption{ \small Components of Partial Observations $o_t$ and Hidden States $s_t$}
    \begin{tabular}{lccc}\toprule
    Entry                   & Dimension     & Noise level  & Category\\
    \midrule
    Command                 & 3             & 0            &Observation\\
    Base Angular Velocity   & 3             & 0.3          &Observation\\
    Base Rotation XY        & 2             & 0.09         &Observation\\
    DoF Position            & 18            & 0.075        &Observation\\
    DoF Velocity            & 18            & 2.25         &Observation\\
    DoF Action              & 18            & 0            &Observation\\
    \midrule
    DoF Position            & 18            & -            &Hidden State\\
    DoF Velocity            & 18            & -            &Hidden State\\
    Base Linear Velocity    & 3             & -            &Hidden State\\
    Base Angular Velocity   & 3             & -            &Hidden State\\
    Base Height             & 1             & -            &Hidden State\\
    \bottomrule
    \end{tabular}
    \label{tab:o_s_def}
    \vspace{-4mm}
\end{table}

\section{Methodology}\label{sec:Approach}
\subsection{Motion Re-targeting and Correction}
\tonghe{ 
To achieve realistic robot movement, we utilize human motion data collected through Motion Capture (MoCap) as the reference motion for Noetix N1 robot. 
Specifically, we first identify key points like knees and elbows, and then scale the source motion to match the robot's size~(cf.~Fig~\ref{fig:retarget}). Next, we apply Inverse Kinematics to compute the joint positions. Finally, we play the motion files in simulators to ensure symmetric body movement. 
}

\subsection{Locomotion Learning with Hybrid Internal Model}\label{section:him_il}
We introduce an anthropomorphic locomotion learning framework combining velocity and implicit state estimation with human
gesture supervision. 
Following the tradition of RL, 
we define a series of reward signals specified in table~\ref{tab:task_reward}) to measure how well the robot accomplishes basic locomotion tasks. 

\begin{table}[ht]\centering
    \caption{ \small Task Reward}
    \begin{tabular}{lcc}\toprule
    Reward                   & Equation $r_i$ & Scale $w_i$ \\
    \midrule
    Feet slip                & $\bm{\omega} \cdot I_{c(t)}$ & -0.05 \\
    Feet contact forces      & $\sum_{i} \left(\| \mathbf{F}_i^{\text{contact}} \| - F_{\text{max}} \right)$ & -0.01 \\
    Lin. velocity tracking & $\exp \left\{-4(\mathbf{v}_{xy}^{\text{cmd}}-\mathbf{v}_{xy})^2\right\}$ & 2.4 \\
    Ang. velocity tracking& $\exp \left\{-4(\bm{\omega}_{\text{yaw}}^{\text{cmd}}-\bm{\omega}_{\text{yaw}})^2\right\}$ & 1.1 \\
    Root accelerations       & $\exp \left\{-\left(\ddot{\mathbf{v}}_{\text{root}}\right)^3\right\}$ & 0.2 \\
    Smoothness               & $(\mathbf{a}_t-2 \mathbf{a}_{t-1}+\mathbf{a}_{t-2})^2$ & -0.01 \\
    \bottomrule
    \end{tabular}
    \label{tab:task_reward}
\end{table}

\begin{figure*}[htp]
    \centering
    \includegraphics[width=0.90\textwidth,height=0.38\textwidth]{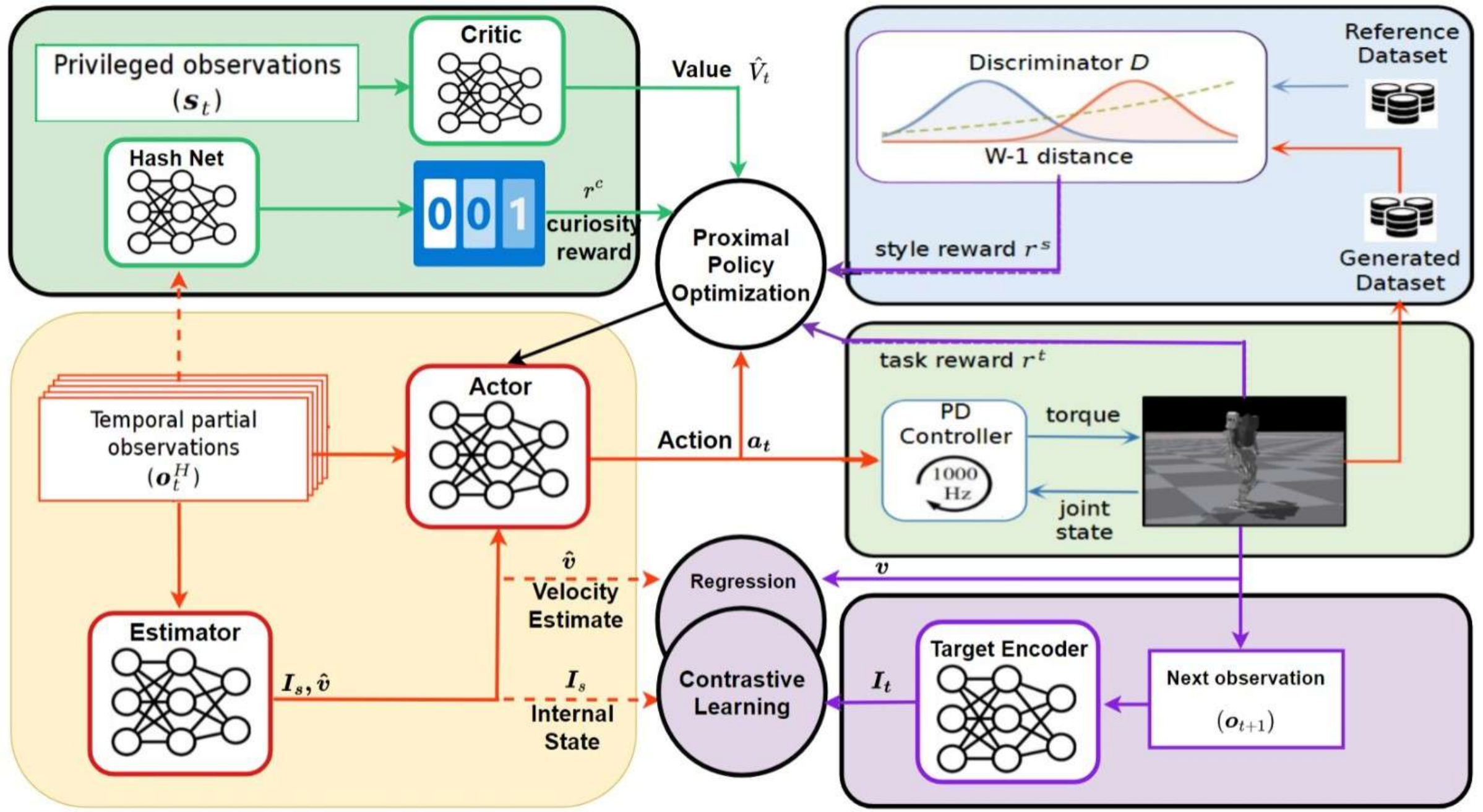}
    \caption{ \small 
    Illustration of the proposed human-like locomotion learning framework. 
        The estimator extracts information from past observations, producing a velocity estimate with internal state representation. The velocity estimate ensures mobile stability, while contrastive learning promotes future observation prediction. Imitation learning secures human-like gaits, and a curiosity bonus fosters exploration. The bottom-left block is employed in real-world deployment.
        }
    \label{fig:pipeline}
    \vspace{-0.15in}
\end{figure*}

To minimize simulation-to-real gap, we apply domain randomization to robot's hardware status and the environment's mechanical properties
according to the configuration described in Table~\ref{tab:dom_rand}. 
\begin{table}[ht]
    \centering
    \small 
    \caption{ \small Domain Randomization}
    \begin{tabular}{p{3.5cm}p{2cm}p{1cm}} 
    \toprule
    Parameter & Range  & Unit \\
    \midrule
    Base mass & [-5, 5] & kg \\
    Center of mass shift & [-0.02, 0.02] & m \\
    Friction coefficient & [0.1, 2] & - \\
    $K_p$ factor & [0.8, 1.2] & N$\cdot$m/rad \\
    $K_d$ factor & [0.8, 1.2] & N$\cdot$m$\cdot$s/rad \\
    Push Force & [-0.6, 0.6] & m/s \\ 
    Push Torque & [-0.6, 0.6] & rad/s \\ 
    Motor strength & [0.8, 1.2] & N$\cdot$m \\
    System delay & [0, 60] & ms \\
    \bottomrule
    \end{tabular}
    \label{tab:dom_rand}
    \end{table}
To take advantage of the anthropomorphism of Imitation Learning while enhancing the adaptability and 
stability of locomotions, we make significant improvements to classical RL framework, which is illustrated in Fig~\ref{fig:pipeline}.
In what follows, we highlight several key designs:
\begin{itemize}
    \item A GAN-based discriminator. It provides additional supervision on the style of the robot's movements. 
    \item A Hybrid Internal Model~\cite{long2024hybridinternalmodellearning} with velocity estimate. 
    It constructs velocity estimate with latent representation of the hidden states from historical observations. 
\end{itemize}

The discriminator serves as an Imitation Learning module that provides a style reward based on the similarity between generated and expert motions, correcting unnatural joint positions. The Hybrid Internal Model (HIM) enhances locomotion imitation flexibility by generating velocity estimates and hidden states, offering rich insights into the relationship between input commands and the robot's status. This method enables humanoids trained with imitation learning and HIM to adaptively learn human-like movement in response to varying commands and diverse environments, thus improving motion stability in real-world scenarios. 

\tonghe{ 
We train the the velocity estimator via direct supervision in the simulator, while we learn the latent state estimates through contrastive learning, encouraging the agent to differentiate command status before making decisions. 
The policy and value networks are trained using Proximal Policy Optimization (PPO). 
}

Since reference motion files include a wide variety of actions such as walking and running, the agent must learn to infer intermediate states for seamless motion transfer, which contrasts with classical Imitation Learning that focuses on mimicking a limited action set. To enhance similarity with expert demonstrations while maintaining generalization, we introduce two modifications to our training pipeline, detailed in Sections~\ref{section:explore} and \ref{section:wgan}.


\subsection{Prevent Mode Collapse by Wasserstein divergence}\label{section:wgan} 

\begin{figure}[htp]
    \centering
        \centering
        \vspace{-1mm}
        \includegraphics[width=0.45\textwidth, height=4.5cm]{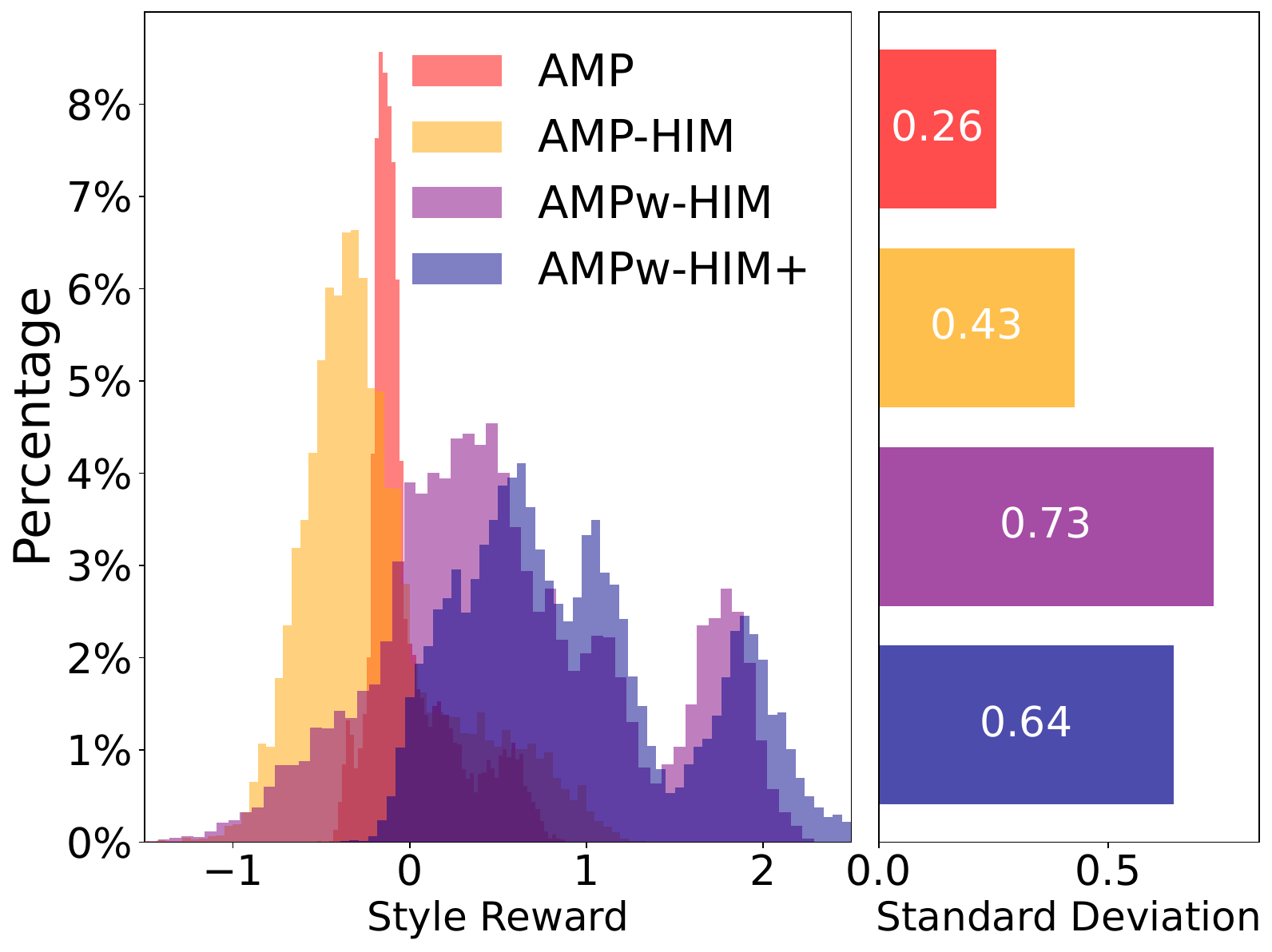}
        \caption{ \small
            Statistics of the style reward $r^{s}$ obtained during training indicate the diversity of joint positions in the output actions. 
            The RL module HIM aids in capturing a broader reward distribution (Left) with a larger standard deviation (Right). 
            Wasserstein divergence (AMPw) captures multiple peaks, 
            a feature not provided by the vanilla AMP design.
            Also note that while curiosity bonus reduces deviation, 
            it shifts the distribution towards a higher-reward region.
        }
        \label{fig:style}
\end{figure}

\tonghe{ 
Our Imitation Learning (IL) module is based on Generative Adversarial Networks (GAN). We train a discriminator network $D(\cdot)$ against a generator $G(\cdot)$ to minimize the difference between real and synthetic data according to a specific criterion. 
To enable diverse motion learning, it is crucial that the IL module produces varied robot motions, which is essential for learning versatile movements through reinforcement learning (RL). 
}
However, traditional GANs and many of their variants face ``mode collapse'' where generators only capture singular peaks in the expert distribution, leading to monotonous outputs (illustrated in Fig~\ref{fig:style}). This limitation may restrict our humanoids to limited joint positions, resulting in stiff motions and reduced tracking ability. 

\tonghe{ 
To make the output of the IL module more diverse, we employ a Wasserstein GAN with divergence (WGAN-div) loss. }
WGAN-div offers a more continuous loss function than original GANs, ensuring that even when actions differ significantly from expert demonstrations, the gradients remain smooth. This property allows for stable training throughout a broader state-action space, ultimately leading to more diverse action distributions.

\subsection{Encourage Exploration via a Curiosity Bonus}\label{section:explore}
Training robots to acquire diverse skills requires 
\tonghe{ 
the agent to explore 
} 
various joint angles. 
\tonghe{ It is difficult to fully describe this diversity with simple, pre-defined reward functions, so we promote the agent to autonomously search for unseen state-action space. Concretely speaking, in addition to the task reward $r$, 
}
we incorporate a curiosity reward $r^c$ into the value function:
$
V := \mathbb{E} \left[ 
\sum_{h=0}^{+\infty} \gamma^t (r_h + r^c_h)
\right]. 
$ 
This curiosity reward $r^c_h(s,a)$ 
\tonghe{ 
is inversely related with the number of historical visits to the state-action pair $(s,a)$ at step $h$.  
}
Since RL maximizes the value function, adding the curiosity term encourages the policy to explore less-visited areas, thus promoting exploration.

\tonghe{
We implement this idea with Curiosity Hash-net~\cite{tang2017explorationstudycountbasedexploration} 
Specifically, we use a neural network 
$\phi$ to map each continuous state $s$ to a binary string, 
which we refer to as the Hashing value. 
To approximate the visitation frequency to the continuous states, we count the number of occurrences of their Hashing values, which gives the definition of the curiosity reward 
$
r^c_t(s_t) := \frac{1}{\sqrt{N(\phi(s_t))}}. 
$
}

\section{Experiments}
\vspace{-1mm}
In this section, we present extensive ablation experiments to quantitatively support the effectiveness of 
the methods proposed in Sections~\ref{section:him_il}, \ref{section:wgan} and \ref{section:explore}.
We center on showing how the HIM module, WGAN-div, and curiosity rewards help shape traditional IL algorithm
(AMP) into a command-adaptive robot learning pipeline with strong generalization ability. 

Section~\ref{section:simulation_exp} is devoted to simulation experiments, while Section \ref{sec: real} demonstrates real-world performance. 
In the following sections, we denote by \textbf{AMP} the baseline agent trained without the RL module.  
It demonstrates the strong capability of IL in precisely mimicking movements, as well as the blatant limitation of learning to adapt to varying commands. 
\textbf{AMP-HIM} is the group trained with IL and HIM module simultaneously, 
while \textbf{AMPw-HIM} replaces the loss function of AMP-HIM from mean-square error to Wasserstein-divergence. 
Adding curiosity rewards to AMPw-HIM, 
we arrive at our full-stack approach, denoted as \textbf{AMPw-HIM+}.
\footnote{For completeness, we also include \textbf{AMPw}, whose agents are trained only with Imitation Learning with WGAN-div.
    }

\subsection{Simulation Experiments}\label{section:simulation_exp}
We evaluate our proposed method in simulations based on the following four aspects, which are crucial yet distinct dimensions for assessing the vividness and effectiveness of a humanoid's movements. 
\begin{enumerate}
    \item \textbf{Basic Locomotion}:
    Evaluates how well the agents accomplish stable and agile locomotions that fit their hardware structure. 
    It is reflected by task rewards such as body collision and feet slippery. 
    \item \textbf{Anthropomorphism}: Measures how similar the robot's gaits are to human data, which 
    adds the vividness and naturalness to basic locomotion.
    \item \textbf{Velocity Tracking}: Assesses how accurately the gait follows the required speeds, 
    which represents mobile adaptability to changing commands. 
    \item \textbf{Generalization}: Tests whether the algorithm is able to generalize to different robot structures, 
    various simulated and real environments and perform unseen motions to complete multiple tasks.
\end{enumerate}

\subsubsection{Basic Locomotion}\label{section:basic_loco}
\begin{figure}[h]
    \centering
    \includegraphics[width=0.45\textwidth,height=0.32\textwidth]{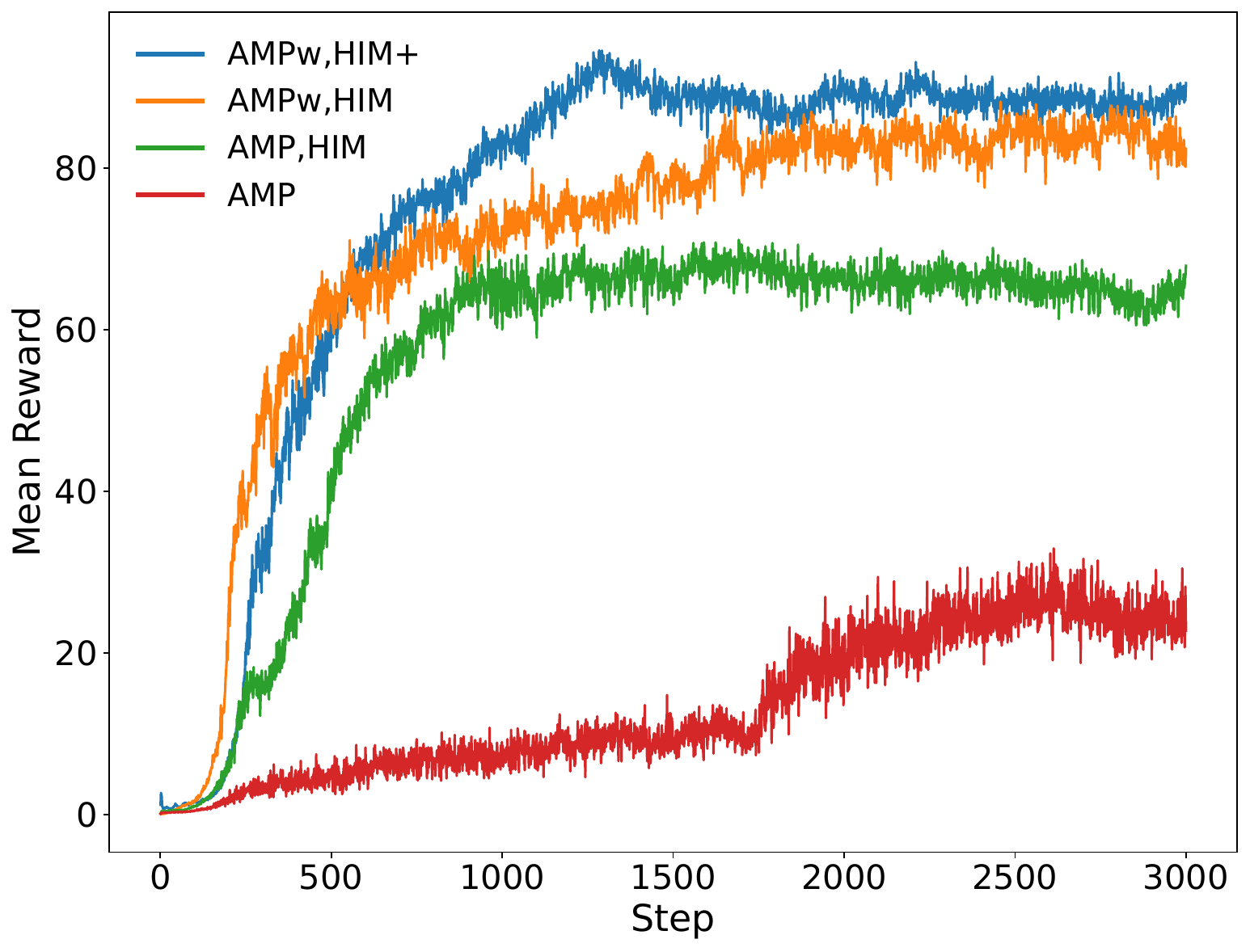}
    \caption{ \small
        Training curves of four algorithms. Vanilla AMP~(red curve) fails to 
    obtain satisfactory task rewards due to its limited generalization. 
    RL-aided methods consistently outperforms AMP baseline.}
    \label{fig:mean_reward_curves}
\end{figure}

We train four groups of agents with the same reward and domain randomization levels 
and record the average task rewards in Fig~\ref{fig:mean_reward_curves}. 

\begin{figure}[h]
    \centering
    \includegraphics[width=0.47\textwidth]{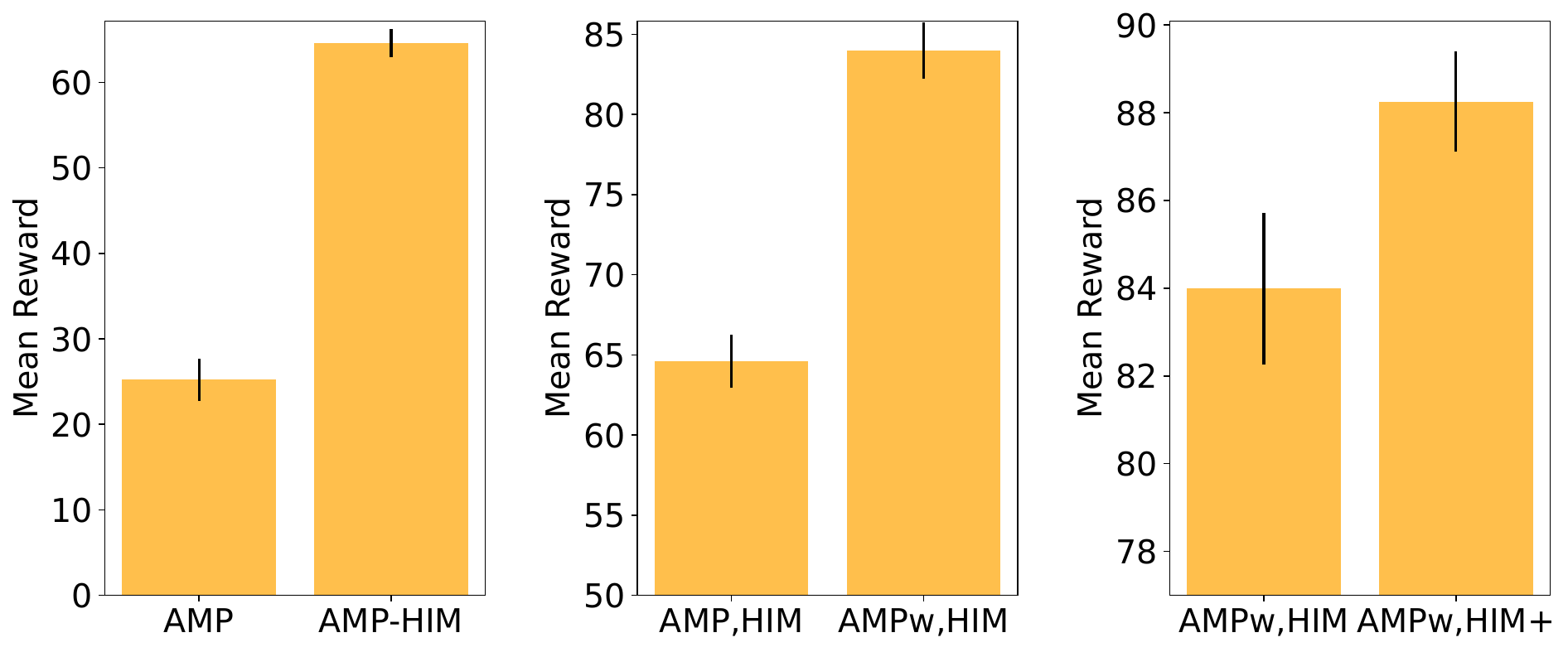}
    \caption{ \small
        Ablation study on the mean reward. 
    Orange bars are the mean returns calculated in the last 500 iters when optimizer converges,  
    error bars indicate their standard deviation.
    HIM module, WGAN-div criterion, and curiosity rewards significantly outperforms 
    the counterparts with these structures removed.
    }
    \label{fig:mean_reward_compare}
    \vspace{-3mm}
\end{figure}

Ablation tests shown in Fig~\ref{fig:mean_reward_compare} 
further highlight the advantages of WGAN-div and curiosity rewards in adapting human motion patterns to the robot's form. 
Compared to \textbf{AMP,HIM}, WGAN produces a greater variety of joint angles for the agent to explore, 
while curiosity rewards help the RL policy escape local minima, 
optimizing within the newly explored state-action space. 
Thus, \textbf{AMPw,HIM} and \textbf{AMPw,HIM+} achieve significant increases in mean returns.

\subsubsection{Anthropomorphism}\label{section:anthropomorphism}
To measure the human-likeness of the generated motions, first we apply different commands to drive the agents to walk, jog, spin and walk backwards, 
capturing motion frames in 20 parallel experiments for each group. Then we compare the generated motions with expert data corresponding to the four movement types, 
calculating their similarity in joint positions using the Dynamic Time Warping (DTW) toolbox~\cite{li2023learning} in Isaac Gym. 
The results are presented in Table~\ref{tab:human-like}. 
The baseline model, AMP, achieved the lowest DTW distances across all four movement categories, 
while the incorporation of HIM hurts human-like performance. This is not surprising, as RL solutions make adjustments to human gestures for better adaptability to the robot 
entity. However, we note that the introduction of WGAN and curiosity mitigates the negative impacts of HIM alone, correcting its joint positions to preserve human's flavor 
captured by the IL module. This correction effect is also evidenced by Figure~\ref{fig:dora} and \ref{fig:real}. 
\begin{table}[t]
    \centering
    \small
    \resizebox{\columnwidth}{!}{%
    \begin{tabular}{lcccc}
      \toprule
      & \multicolumn{4}{c}{DTW Distance} \\
      \cmidrule(lr){2-5}
      Method & Walking & Jogging & Walking Backwards & Turning Around\\
      \midrule
        AMP         & 1160.53 $\pm$ 21.68           & 1488.90 $\pm$ 30.16           & 736.70 $\pm$ 4.02             & 990.19 $\pm$ 8.53 \\
        AMP,HIM     & 1417.92 $\pm$ 20.13           & 1638.21 $\pm$ 21.17           &755.56 $\pm$ 2.07              & 1075.36 $\pm$ 4.06 \\
        AMPw,HIM    & \textbf{1402.52 $\pm$ 23.55}  & 1640.28 $\pm$ 18.52           & 756.66 $\pm$ 1.81             & 1080.10 $\pm$ 4.68 \\
        AMPw,HIM+   & 1404.08 $\pm$ 20.32           & \textbf{1637.08 $\pm$ 15.75}  & \textbf{742.49 $\pm$ 1.65}    & \textbf{1054.76 $\pm$ 3.64} \\ \midrule
    \end{tabular}%
    }
    \caption{ \small
        Comparison of anthropomorphism between different strategies. A lower DTW score 
        indicates stronger similarity with human expert data. 
        Bold figures indicate the best performance in HIM-aided groups, 
        which shows that WGAN-div and curiosity helps preserve human flavor in various 
        locomotion tasks.
    }
    \label{tab:human-like}
\end{table}

\subsubsection{Velocity Tracking}
\begin{figure}[h]
    \centering
    \includegraphics[width=0.48\textwidth,height=0.30\textwidth]{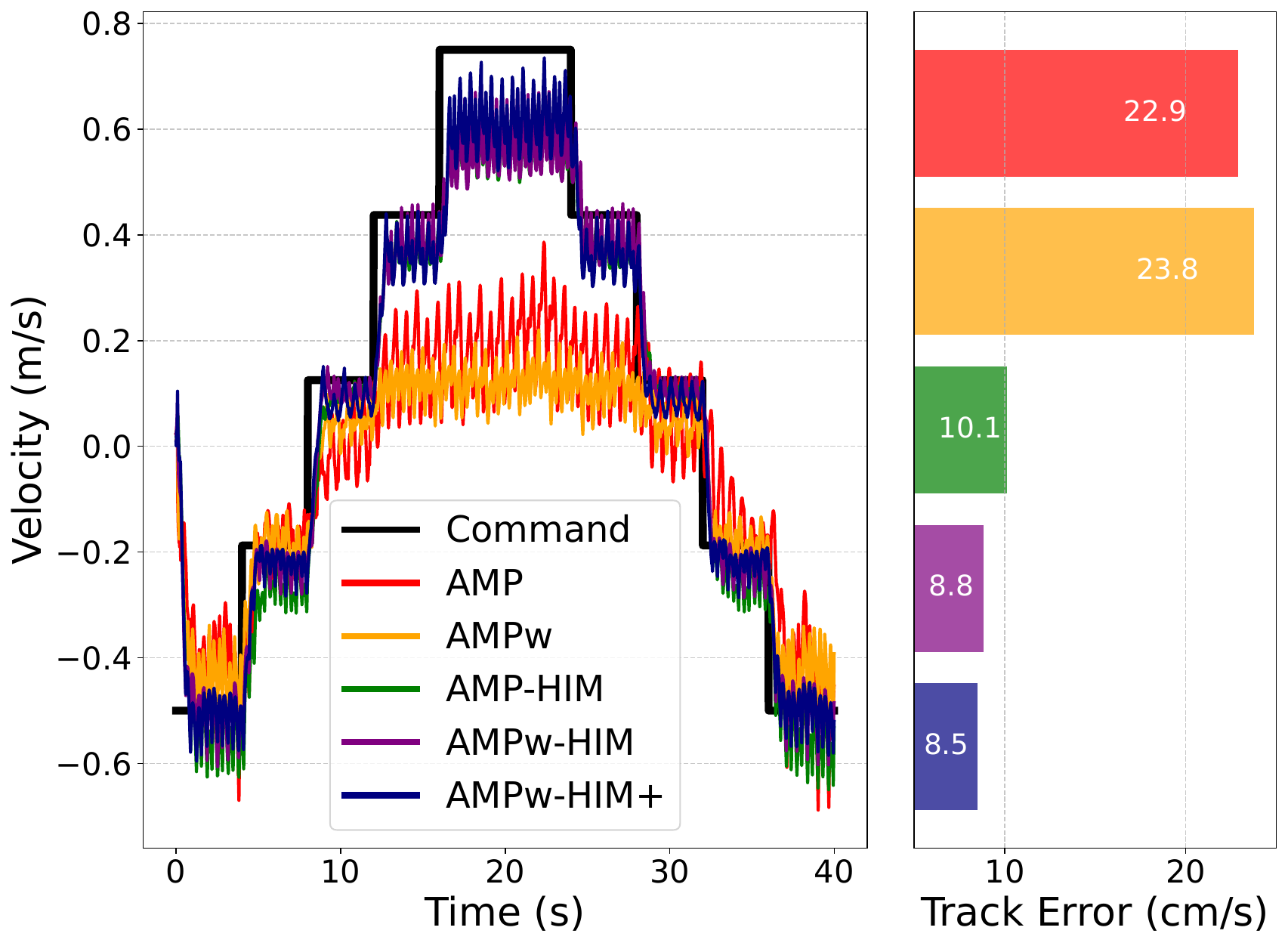}
    \caption{ \small 
            Comparison of different algorithm's velocity tracking ability. 
            Applied with abruptly changing velocities commands, 
            pure IL methods are unable to follow unseen velocity requirements~(Left), while 
            HIM-aided policies do, yielding significantly smaller tracking error~(Right). 
        }
        \label{fig:vel_track}
\end{figure}

Apart from comparing the ability to mimic separate human motions, we also test 
how the humanoids adapt to varying commanded velocities in Isaac Gym, whose results are shown in 
Fig~\ref{fig:vel_track}. 
The speed region of the reference motion files is concentrated within $[-0.4m/s,0.4m/s]$. 
We apply velocity commands that sweep within $[-0.5m/s,0.75m/s$], making abrupt leaps during acceleration and 
deceleration. \textbf{AMP}-based methods output actions that stuck the robot within a small velocity region, 
while RL-aided groups such as \textbf{AMP,HIM} exhibit a significant increase in velocity tracking ability. 
To show that HIM alone is able to boost performance for various IL methods, we add an additional experiment with group \textbf{AMPw}, 
who follows commands even poorer than the AMP baseline, standing in sharp contrast with \textbf{AMPw,HIM}. 
Ablation studies in Fig~\ref{fig:mean_reward_compare} strongly prove that HIM greatly and consistently enhances the mobile adaptability upon different imitation learning methods. 
The success of HIM comes from its velocity and latent state estimators.

\subsubsection{Generalization}
Equipped with HIM and curiosity rewards, our method fosters strong generalization ability in various dimensions. 

\begin{figure}[h]
    \vspace{-6mm}
    \centering
    \begin{subfigure}[t]{0.1\textwidth}
        \centering
        \includegraphics[height=2.5cm]{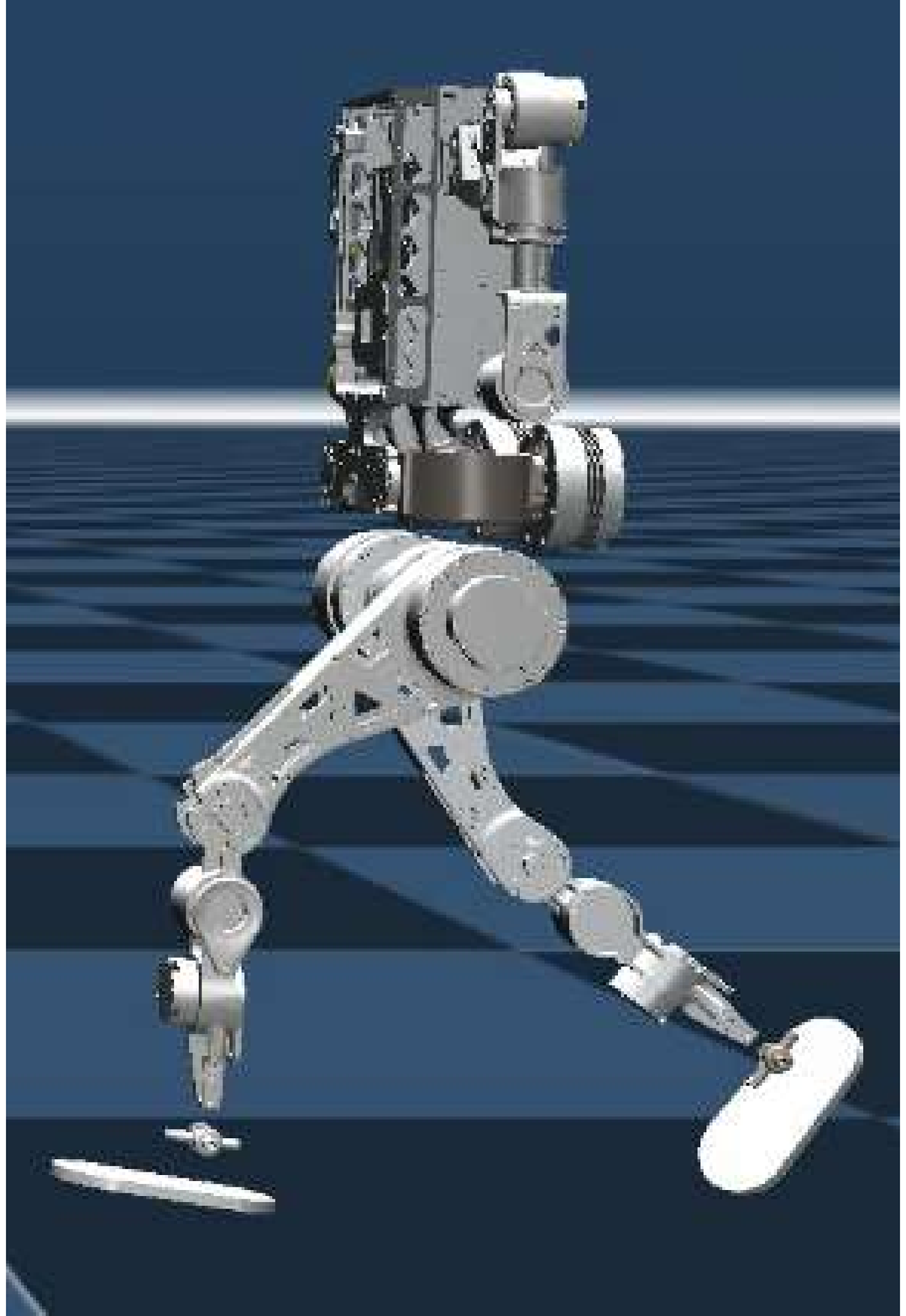}
        \caption{ \small Run}
        \label{fig:running}
    \end{subfigure}
    \hfill
    \begin{subfigure}[t]{0.1\textwidth}
        \centering
        \includegraphics[height=2.5cm]{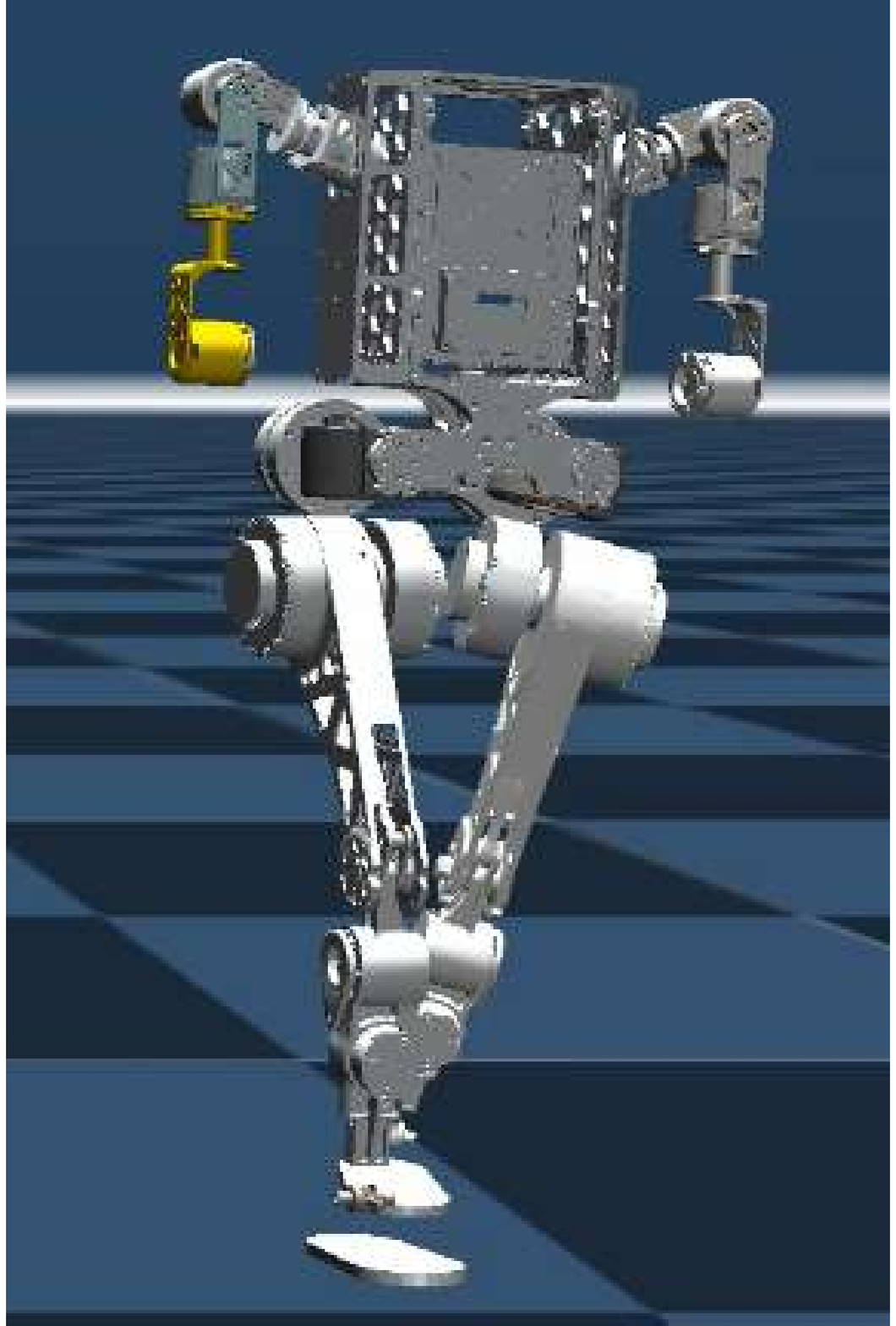}
        \caption{ \small Spin}
        \label{fig:spinning}
    \end{subfigure}%
    \hfill
    \begin{subfigure}[t]{0.12\textwidth}
        \centering
        \includegraphics[height=2.5cm,width=1.9cm]{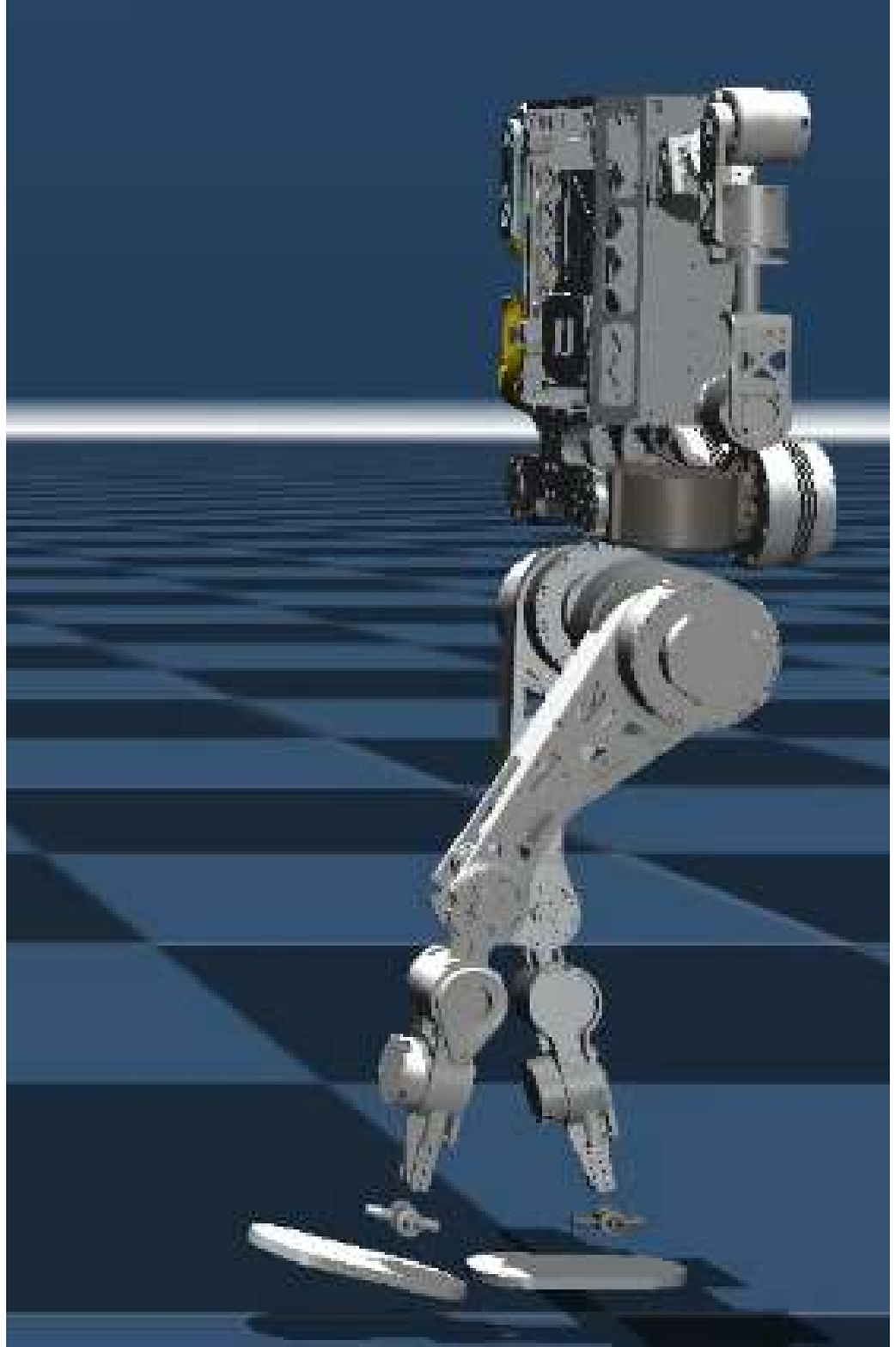}
        \caption{ \small Backward}
        \label{fig:backward}
    \end{subfigure}%
    \hfill
    \begin{subfigure}[t]{0.12\textwidth}
        \centering
        \includegraphics[height=2.5cm,width=2.1cm]{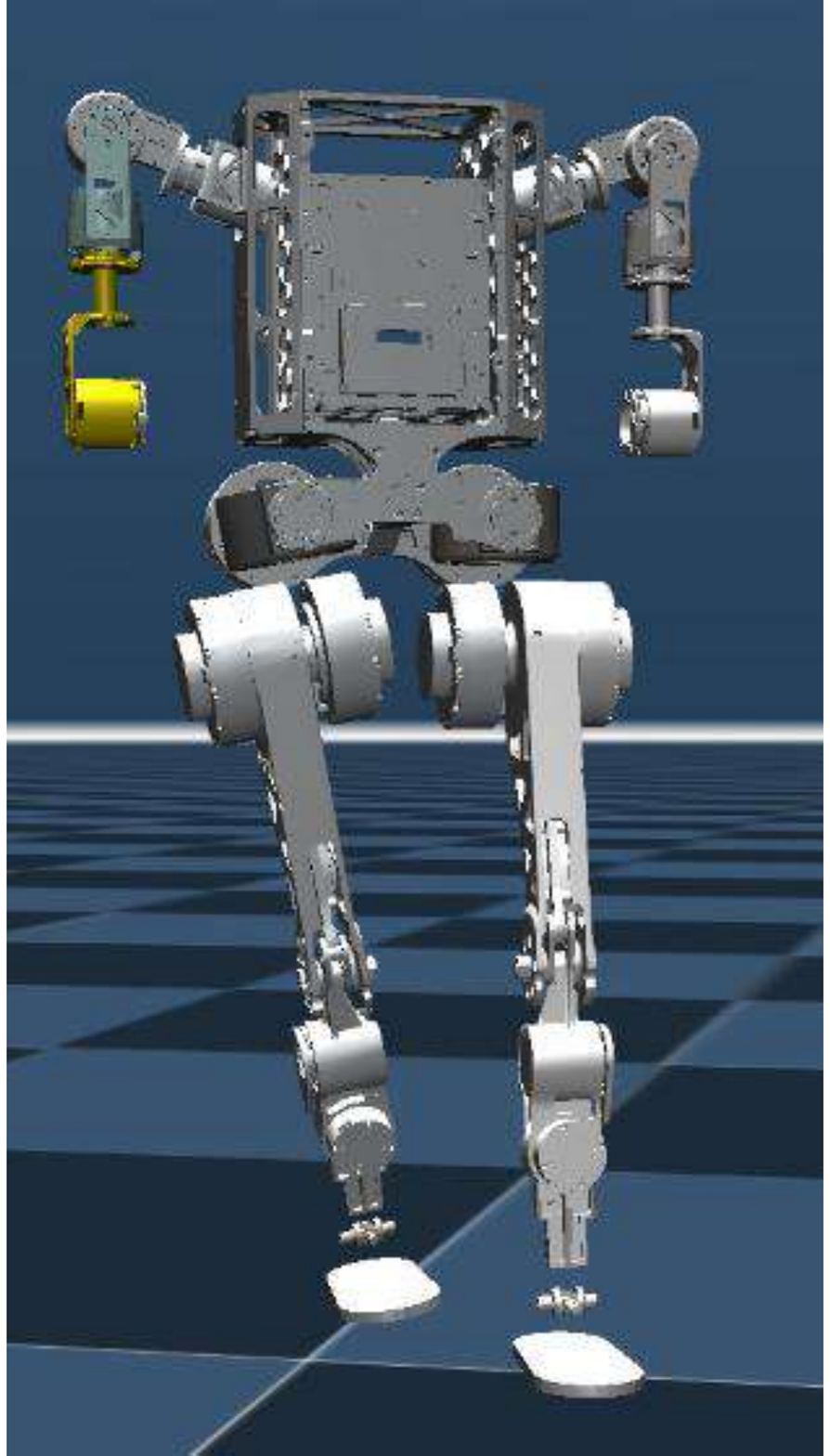}
        \caption{ \small Sideways}
        \label{fig:sideways}
    \end{subfigure}
    \caption{ \small 
    \tonghe{
    Noetix Dora learns body movements different from the reference motion after zero-shot transfer to MuJoCo.
    }
    }
    \label{fig:dora}
    \vspace{-4mm}
\end{figure}

\textbf{a) Hardware adaption.} 
Our method endows different robots with versatile locomotion ability, 
including Noetix N1 in Fig~\ref{fig:real} and Dora in Fig~\ref{fig:dora}. 

\textbf{b) Zero-shot transfer in various environments.}
After training in Isaac Gym, our robot agents are able to smoothly transfer to the other simulation environment MuJoCo, without any additional engineering. 
\tonghe{
}
We can directly employ the same policy to robots in the real world, using exactly the same set of PD controller parameters, which is shown in Fig~\ref{fig:real}.

\textbf{c) Motion generalization and command-adaptive multitasking.} 

We witness direct evidence of motion generalization in the experiments. 
We trained agents only with expert data of straight walking and running, but the policies returned by \textbf{AMPw,HIM} are able to perform various unseen motions desired by various input commands, such as walking backwards, taking sidesteps, and spinning while running~(cf. Fig~\ref{fig:dora}).
\tonghe{
As an explanation}, Fig~\ref{fig:style} shows WGAN-div captures the expert distribution with multiple peaks, indicating the generated joint positions possess stronger variety. This further implies RL-boosted IL is able to break expert motion sequences down to basic joint position combinations with much finer granularity, thus helping agents assemble basic action units to form complex action suites, resulting in novel motions.

\begin{figure}[h]
    \centering
    \includegraphics[width=0.30\textwidth,height=0.20\textwidth]{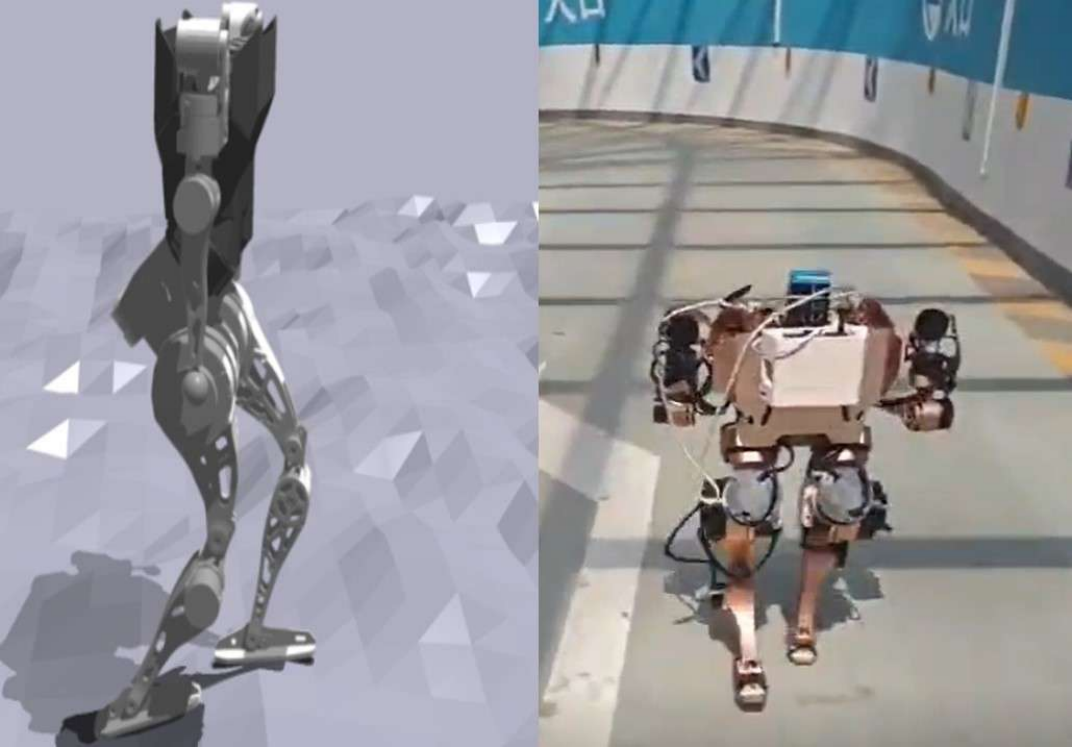}
    \caption{ \small 
    \tonghe{
    Noetix N1 harnesses complex terrains, including bumpy surfaces in Isaac Gym (Left) and 
    descending slopes in the real world (Right).
    }
    }
    \label{fig:ning_terrain}
    \vspace{-5mm}
\end{figure}

\textbf{d) Harnessing complex terrains.} 
The HIM module distinguished different motion files and terrains, making our algorithm applicable to complex terrains like bumpy surfaces and descending planes, as shown in Fig~\ref{fig:ning_terrain}. 

\tonghe{
}


\subsection{Real-world Evaluation}\label{sec: real}

\begin{figure}[h]
    \centering
    \includegraphics[width=0.48\textwidth,height=0.37\textwidth]{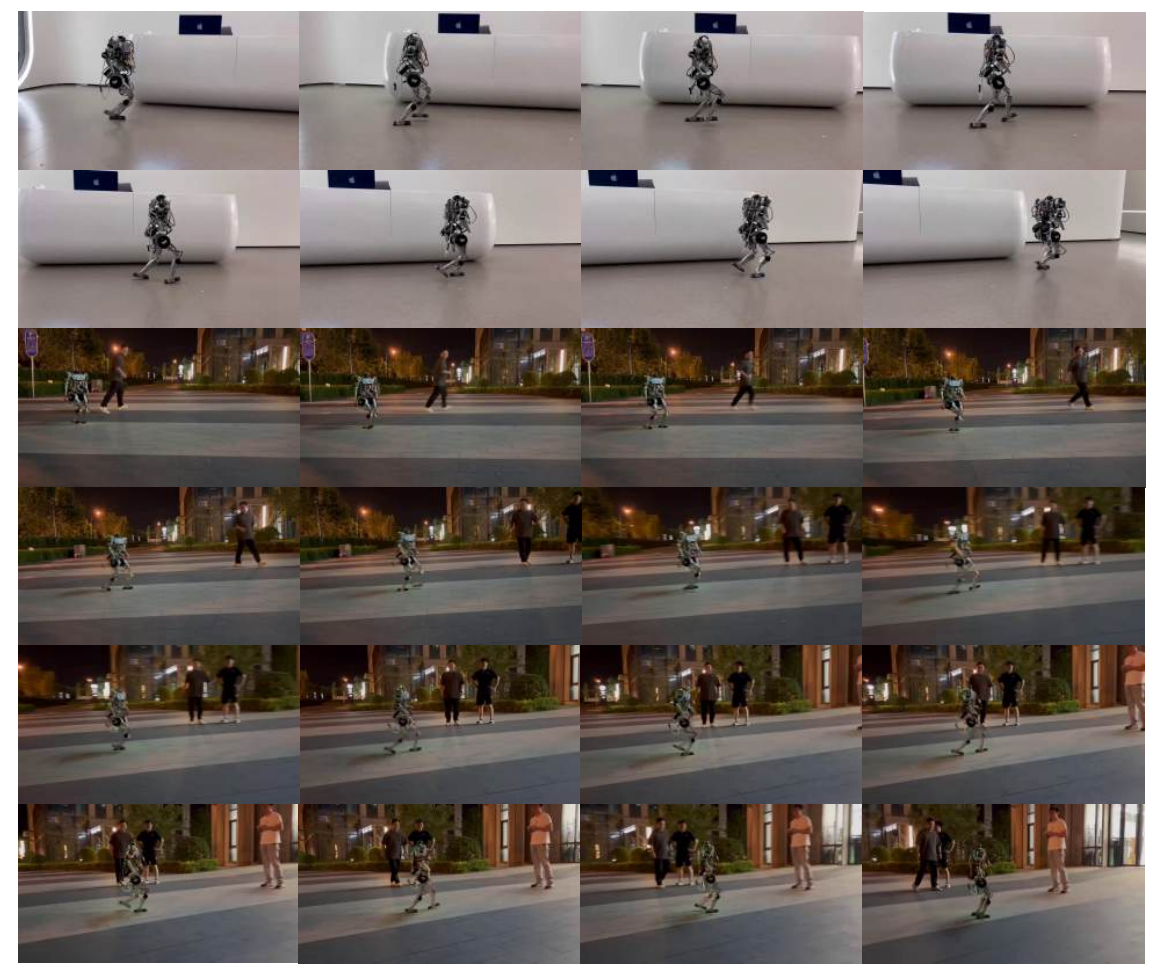}
    \caption{ \small 
    \tonghe{
    Noetix N1 walks and runs walks and runs stably walking and running ability in the real world, both indoors and outdoors. 
    }
    }
    \label{fig:real_experiment}
    \vspace{-5mm}
\end{figure}

We also conduct extensive real-world experiments for robot Noetix N1.  
\tonghe{
We remotely control the robot using a joystick while applying varying velocity commands.
}
For example, in Fig.~\ref{fig:real}, the robot's gaits in the real-world highly align with other images in the two simulators, 
which demonstrates the strong robustness and generalization ability of our algorithm. 
Moreover, we show in Fig~\ref{fig:real_experiment} that Noetix N1 is able to walk, run and stop in different environments 
following various command series. 

\section{Conclusion and Future Work}\label{sec:conclusion}
    This work introduces a novel humanoid locomotion learning framework that smoothly transitions between human-like motions based on changing commands. By using a curiosity bonus and the WGAN-divergence criterion, our method enhances the AMP algorithm's generalization, while a hybrid internal model simultaneously tracks velocity and estimates unobserved states, significantly improving adaptability. Comprehensive simulations and real-world experiments validate the effectiveness of our approach.

    We can extend existing findings in several directions, 
    which include adapting our method to diverse terrains such as stairs and slopes
    and enhancing the versatility of robot motions by learning to hop or coordinate with hand manipulation tasks.


\bibliographystyle{./IEEEtranS.bst}
\bibliography{ref}

\end{document}